\documentclass{article}

\usepackage{arxiv}

\usepackage[utf8]{inputenc} 
\usepackage[T1]{fontenc}    
\usepackage{hyperref}       
\usepackage{url}            
\usepackage{booktabs}       
\usepackage{amsfonts}       
\usepackage{nicefrac}       
\usepackage{microtype}      
\usepackage{amsmath}
\usepackage{cleveref}       
\usepackage{lipsum}         
\usepackage{graphicx}
\usepackage{natbib}
\usepackage{doi}
\usepackage{authblk}

\usepackage[T1]{fontenc}
\usepackage{graphicx}
\usepackage{booktabs}
\usepackage[misc]{ifsym}

\usepackage{graphicx}
\usepackage{blindtext}
\usepackage{glossaries}
\usepackage{mathtools}
\usepackage{wrapfig}
\usepackage{algorithm}
\usepackage[noend]{algpseudocode}
\usepackage{multirow}
\usepackage{subfig}
\usepackage[table, svgnames, dvipsnames]{xcolor}
\usepackage{makecell, cellspace, caption}

\newacronym{CNF}{CNF}{Conjuctive Normal Form}
\newacronym{ARM}{ARM}{Association Rule Mining}
\newacronym{NARM}{NARM}{Numerical Association Rule Mining}
\newacronym{ARL}{ARL}{Association Rule Learning}
\newacronym{RDF}{RDF}{Resource Description Framework}

\title{AE SemRL: Learning Semantic Association Rules with Autoencoders}

\newbox{\orcid}\sbox{\orcid}{\includegraphics[scale=0.06]{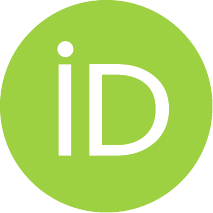}} 
\author[1]{%
	\href{https://orcid.org/0000-0003-2710-7951}{\usebox{\orcid}\hspace{1mm}Erkan Karabulut}%
}
\author[1]{%
	\href{https://orcid.org/0000-0001-7054-3770}{\usebox{\orcid}\hspace{1mm}Victoria Degeler}%
}
\author[1]{%
	\href{https://orcid.org/0000-0003-0183-6910}{\usebox{\orcid}\hspace{1mm}Paul Groth}
}
\affil[1]{University of Amsterdam, 1098 XH, North Holland, The Netherlands \protect\\ \texttt{\{e.karabulut, v.o.degeler, p.t.groth\}@uva.nl}}

\renewcommand{\headeright}{}
\renewcommand{\undertitle}{}
\renewcommand{\shorttitle}{AE SemRL: Learning Semantic Association Rules with Autoencoders}

\hypersetup{
pdftitle={AE SemRL: Learning Semantic Association Rules with Autoencoders},
pdfauthor={David S.~Hippocampus, Elias D.~Striatum},
pdfkeywords={Rule learning, Association rules, Autoencoders, Semantics, Time series data},
}

\begin{document}
\maketitle

\begin{abstract}
\gls{ARM} is the task of learning associations among data features in the form of logical rules. Mining association rules from high-dimensional numerical data, for example, time series data from a large number of sensors in a smart environment, is a computationally intensive task. In this study, we propose an Autoencoder-based approach to learn and extract association rules from time series data (AE SemRL). Moreover, we argue that in the presence of semantic information related to time series data sources, semantics can facilitate learning generalizable and explainable association rules. Despite enriching time series data with additional semantic features, AE SemRL makes learning association rules from high-dimensional data feasible. Our experiments show that semantic association rules can be extracted from a latent representation created by an Autoencoder and this method has in the order of hundreds of times faster execution time than state-of-the-art ARM approaches in many scenarios. We believe that this study advances a new way of extracting associations from representations and has the potential to inspire more research in this field.
\end{abstract}


\section{Introduction}

Association Rule Mining (ARM) aims to discover commonalities in data in the form of logical rules and has successful applications in a wide variety of domains and tasks such as anomaly detection~\citep{nassif2021machine}, recommendation systems~\citep{khanal2020systematic}, and product design~\citep{jiang2018multi}. Initial ARM algorithms~\citep{agrawal1994fast,han2000mining} were designed for categorical data in the form of transactions. However, when applied to numerical data with greater size and dimension, ARM algorithms are inefficient to execute~\citep{kaushik2023numerical,telikani2020survey}. Despite the broad range of success of deep learning~\citep{sarker2021machine}, including learning logical rules from graph data~\citep{qu2020rnnlogic}, applying deep learning-based methods directly for learning association rules in a transaction dataset is not yet well-studied.


Besides inefficiency, ARM also faces the challenge of producing a high number of rules that are difficult to explain~\citep{kaushik2023numerical}. Processing a large number of rules e.g., to decide which ones are useful for a certain task is computationally expensive and time-consuming. Explainability is particularly important when it comes to verifying the rules by a human expert or an automated symbolic system. This paper focuses on addressing these three issues in the case of learning association rules from mixed categorical and numerical time series data in the form of transactions.


To overcome the challenges of rule quantity and explainability, we utilize semantics. Knowledge graphs \citep{kg-book} - large database of structured semantic information - are being increasingly used with time series data e.g., from sensors in a smart environment~\citep{karabulut2023ontologies} or video processing~\citep{greco2021use}. When knowledge graphs are linked to time series data, association rules with semantic properties can be learned facilitating generalizability and explainability. For instance, in a smart environment without semantics, an association rule might state: \textit{`if sensor1 measures a value in range R, then sensor2 must measure a value in range R2'}. With semantics, rules can take a more contextual form, as seen in the water network domain example: \textit{if a sensor with type T placed inside a Pipe P measures a value in range R, then another sensor with type T2 inside a Junction J that is connected to P must measure a value in range R2}. The association rule with semantics is no longer about individual time series data sources (sensor1 and sensor2). Instead, it describes a certain context that the time series data source is in, and therefore is more generically applicable and explainable. 

On the other hand, enriching time series data with semantics increases input size and therefore results in even higher execution times. We show that the state-of-the-art ARM algorithms have impractical execution times when run on a realistic semantically enriched time series dataset. Therefore, there is a greater need to apply machine learning-based solutions when utilizing semantics.

Hence, this paper proposes an Autoencoder-based approach (AE SemRL) to learn semantic association rules. We show that Autoencoders can learn associations among their input features, and the associations can be extracted from their latent representation in the form of logical rules. We evaluate our approach on three time series datasets with associated semantics from two domains; water networks, and energy. Our experiments show that AE SemRL has hundreds of times faster execution time than the state-of-the-art methods in many scenarios, and the association rules learned by AE SemRL imply strong associations based on commonly used rule quality criteria. 


A list of concrete contributions of this paper is given below:

\begin{itemize}
    \item In Section \ref{sec:problem-statement}, we provide a formal problem definition for semantic association rule learning from time series data and knowledge graphs.
    \item Section \ref{sec:ae-semrl} introduces the first deep learning-based approach (AE SemRL) to learn semantic association rules from time series data. 
    \item Section \ref{sec:experiments} provides empirical evidence that a latent representation created by an Autoencoder captures associations among its input features and the associations can be extracted in the form of logical rules, which makes association rule learning feasible with high-dimensional data. 
\end{itemize}

Section \ref{sec:related-work} presents the background concepts and related work. Section \ref{sec:discussion} discusses current and future perspectives for learning semantic association rules. Lastly, Section \ref{sec:conclusion} concludes with a summary of findings and future work. 

\section{Related Work} \label{sec:related-work}

This section introduces the related work and background concepts.

\subsection{\gls{NARM}} \label{sec:NARM}

Association rules are first-order horn clauses which are formal logical statements in \gls{CNF} with at most one positive literal. They are in the form of an implication, e.g., $X \rightarrow Y$ which is read as \textit{if X then Y}. X is called the \textit{antecedent} or \textit{body}, and Y is called \textit{consequent} or \textit{head} of the rule. 

Following the categorization in a recent extensive systematic literature review by Kaushik et al.~\citeyearpar{kaushik2023numerical}, there are 3 main categories of approaches to NARM: discretization, optimization, and statistical methods. In the discretization-based methods, numerical data is divided into intervals in a way that can facilitate the mining of interesting association rules. It is further categorized into partitioning, clustering, fuzzy, and hybrid approaches. However discretization-based algorithms may suffer from information loss. Another major category of NARM is optimization-based algorithms that include evolutionary, differential evolution, swarm intelligence, and physics-based approaches. These algorithms essentially provide a heuristic-based search process that aims to find association rules that optimize one or more (multi-objective) rule quality criteria. Therefore they are named optimization-based algorithms. These methods as well as the statistical methods that do not include learning a model of input data suffer from handling big high-dimensional data, together with other broader issues in NARM such as having a large number of rules, and explainability as also mentioned by Kaushik et al.~\citeyearpar{kaushik2023numerical} and others~\citep{berteloot2023association,kishore2021applications}. 


\subsection{Association Rule Mining and Semantics}

There have been a number of works focused on association rule mining over knowledge bases typically using Resource Description Framework (RDF) as the representation language~\citep{barati2017mining}. Data mining from RDF also suffers from the efficiency problem~\citep{d2020machine}. Our approach differs in that we are focused on learning association rules from time series data that has semantic properties.

There has been only one study utilizing semantics when learning association rules from time series data ~\citep{karabulut2023semantic}. This existing approach is based on the well-known FP-Growth~\citep{han2000mining} algorithm and uses knowledge graphs to enrich pre-discretized time series data with semantics. We adopt a similar enrichment approach to that work but develop a completely new model based on deep learning. 


\subsection{Deep Learning-based Association Rule Mining}

Deep learning has been part of rule learning tasks either directly or as part of a pipeline of operations. In a recent example, Qu et al.~\citeyearpar{qu2020rnnlogic}, proposed a rule learning approach for learning logical rules on knowledge graphs. Simsek et al.~\citeyearpar{simsek2021deep} utilized and compared many deep learning architectures as part of a rule extraction task, however only to label unstructured data which is then fed into a rule mining algorithm. Another common usage of deep neural networks for rule learning is to understand the effect of input features on a prediction task~\citep{chakraborty2020rule}. This is different than learning associations as it is an unsupervised task.

Patel et al.~\citeyearpar{patel2022innovative} proposed to use of Autoencoders for learning frequent patterns in a grocery dataset, which are then used to create association rules. However, no source code or pseudo-code is given in the paper. Autoencoders are used to create a lower-dimensional representation of a given input data in the literature~\citep{chen2023auto}. To the best of our knowledge, Berteloot et al.~\citeyearpar{berteloot2023association}, proposed the only deep learning-based approach to learn association rules directly (ARM-AE). ARM-AE is an Autoencoder-based ARM algorithm for categorical tabular datasets only, and not for time series datasets. ARM-AE extracts associations from a trained Autoencoder based on the reconstruction success for a given set of input vectors. It uses an Autoencoder with equal size layers and assumes that input to the trained Autoencoder represents a consequent while the successful reconstruction represents antecedents of an association rule. We argue that this assumption does not hold in real data, and input to an inference step should represent an antecedent while the output represents a consequent. We adapt ARM-AE to the case of semantic association rule learning from time series data. Furthermore, we provide an improved Autoencoder architecture that employs denoising with probability distributions in its output layer per input feature, as well as a new algorithm to extract associations from a trained Autoencoder (AE SemRL). 

\section{Problem Definition} \label{sec:problem-statement}

Association rules are in the form of implications, e.g. $X \rightarrow Y$, where $X \rightarrow Y$ is a horn clause with $|Y| = 1$ referring to a single literal and $|X| \geq 1$ referring to a set of literals. A horn clause is defined as \textit{a disjunction of literals with at most one positive literal}.

Given a time series dataset T mapped to a knowledge graph G with binding B, produce a set of semantic association rules with clauses based on T and G. An example of T would be sensor measurements or frame in a video. Note that the T is converted to a set of transactions before the learning process. G is in the form of a directed property graph which contains semantic information regarding the items in T, such as where a sensor is placed. The third one is a binding B that reports which item in T corresponds to which node in G, assuming that each item has a representation in G. Output rules can express conditions on the sensor measurements and its context such as the environment that the sensor is placed in, or a frame in a video and entities in the frame such as cats and dogs.


\subsection{Input}

\begin{table}[t]
    \centering
    \caption{Input notation, explanations and examples from water networks domain.}
    \label{tab:input-notation}
    \begin{tabular}{p{0.08\textwidth}p{0.41\textwidth}p{0.41\textwidth}}
        \toprule
        Notation & Explanation & Example \\ 
        \midrule
        C & Classes in an Ontology/Data schema & Pipe, Junction \\
        R, r & Relations in between the classes & (Pipe)\_connectedTo\_(Junction) \\
        A, a & Properties for the classes and relations in the Ontology/Data schema & (Junction).elevation, elevation property of the class Junction \\
        V & Node IDs in the knowledge graph & P1, J2 \\
        E, e & IDs of the edges in between nodes in the knowledge graph & (P1)\_(e1)\_(J2), P1 and J2 are node IDs, e1 is an edge ID \\
        L, l & Labels for the nodes and edges in the knowledge graph & (P1:Pipe)\_(e1:connected\_to) \_(J2:Junction) \\
        P, U, p & Property-value pairs for nodes and edges in the knowledge graph & (P1:Pipe).elevation=v1, v1 is the value for the elevation property of P1 which is a Pipe \\
        M, S, s & set of value and source pairs & a sensor with the ID s1, measures u1 \\
        M, F, f & set of values and timestamps & a value v1 is obtained at the t1 \\
        V, S, b & there is a node in the knowledge graph for each source & each sensor is represented with a node in the knowledge graph \\
        \bottomrule
    \end{tabular}
\end{table}

\textbf{Knowledge graph.} The knowledge graph described in this section is a property graph with an ontology or data schema as the underlying structure~\citep{tamavsauskaite2023defining}. We adapt the definition for a \textit{property graph}, given in the next paragraph, from the \textit{Knowledge Graphs} book by Hogan et al.~\citeyearpar{kg-book} (see Table \ref{tab:input-notation} for high-level explanations of the notation and examples): 

\textbf{Property Graph.} Let \textit{Con} be a countably infinite set of constants. A property graph is a tuple $G=(V, E, L, P, U,e,l,p)$, where $V \subseteq Con$ is a set of node IDs, $E \subseteq Con$ is a set of edge IDs, $L \subseteq Con$ is a set of labels, $P \subseteq Con$ is a set of properties, $U \subseteq Con$ is a set of values, $e:E \rightarrow V \times V$ maps an edge ID to a pair of node IDs, $l:V \cup E \rightarrow 2L$ maps a node or edge ID to a set of labels, and $p:V \cup E \rightarrow 2P \times U$ maps a node or edge ID to a set of property–value pairs.

\textbf{Ontology/Data Schema.} Let $O=(C, R, A, r, a)$ be an ontology or data schema, where $C \subseteq Con$ is a set of classes, $R \subseteq Con$ is a set of relations, $A \subseteq Con$ be a set of properties, $r:R \rightarrow C \times C$ maps a relation to a pair of classes, and $a:C \cup R \rightarrow 2P$ maps a class or a relation to a set of properties. 

To express that G has O as its underlying structure, we define the following; i) $L \subseteq C \cup R$, meaning that the labels in G can only be one of the classes or relations defined in O, ii) $P \subseteq A$, meaning that the properties of V and E in G, can only be one of the properties in A. This definition is enough to show that knowledge graphs can help in learning generalizable association rules.

\textbf{Time series data.} We define time series data generically as a tuple $T = (M, S, F, s, f)$, where $M \subseteq Con$ is a set of values, $S \subseteq Con$ is a set of source IDs, $F \subseteq Con$ is a set of timestamps and $s:S \rightarrow M$ where $\exists x, x' \in S, \neg (s(x) = s(x') \implies x = x')$ (non-injective), and $\forall y \in M (\exists x \in S (s(x) = y))$ (surjective). A source ID refers to the source of a value M, such as a sensor ID, or a frame ID in a video. This means that s maps each source ID to one or more values in a way that all values are matched with exactly one source ID. Finally, $f:M \rightarrow F$ maps each value to a timestamp. Note that there is no assumed order for timestamps since the task is not to learn temporal rules, and the time series data is represented as a set of transactions to enable generalizable rule learning. 

\subsubsection{Binding}\label{definition:binding}

Binding is a tuple $B = (V, S, b)$, where $V$ is the set of node IDs from G, and $S$ is the set of sources from T, $b: V \rightarrow S$ maps each source to a node in G, and $S \subseteq V$ meaning that there is a node ID for each source ID, and there can be node IDs for more e.g., instances of classes in C. 

\subsection{Output}\label{definition:output}

The output is a set of association rules of the below-described form.

Let $I=\{i'_1, i'_2, ..., i'_b\}$ be a set of $b$ items. We define the following 5 forms for an item: $\forall i' \in I (((i' = (p' \# z')) \lor (i' = (m' \# z')) \lor (i' = (v'_l = l')) \lor (i' = (e'_l = l')) \lor (i' = (v' \rightarrow v'' = e')) )$, with $p' \in P$, $m' \in M$, $v', v'' \in V$, $e' \in E$, $l', v'_l, e'_l \in L$ where $v'_l$ refers to a label mapped to a node with the ID $v'$, and $e'_l$ refers to a label mapped to an edge with the ID $e'$. $z'$ refers to a value that is either \textit{categorical} or \textit{numerical}, $\#$ refers to one of the comparison operations with a truth value defined below:

$\#_{categorical}(p, g) ::= (p = g) | (p \neq g) | (p \in \{g\}) | (p \notin \{g\}) $

$\#_{numerical}(p, g) ::= (p = g) | (p \neq g) | (p > g) | (p < g) | (p \leq g) | (p \geq g)$

\begin{table}[htp]
    \centering
    \caption{Item forms, explanations and examples from water network domain.}
    \label{tab:output-examples}
    \begin{tabular}{p{0.2\textwidth}p{0.25\textwidth}p{0.4\textwidth}}
        \toprule
        Form & Example & Explanation \\ 
        \midrule
        $(i' = (p' \# z'))$ & $(p1:).length > 100$ & A node p1 has length bigger than 100 \\
        $(i' = (m' \# z'))$ & $(s1:Sensor).value < 10$ & A sensor s1 with measurement value smaller than 10 \\
        $(i' = (v'_l = l'))$ & $(p1:Pipe)$ & A node p1 has the label 'Pipe' \\
        $(i' = (e'_l = l'))$ & $(e1:Junction)$ & An edge e1 has the label 'Junction' \\
        $(i' = (v' \rightarrow v'' = e'))$ & $(p1:) \rightarrow (p2:) = (e1:)$ & node p1 is connected to p2 with the edge e1 \\
        \bottomrule
    \end{tabular}
\end{table}

$X \rightarrow Y$ is an implication (association rule) where $(X, Y \subseteq I) \wedge (|Y| = 1)$. This means that on both sides of the implication, items can only consist of properties of classes or relations defined in the ontology, and the consequence side of the implication can only have 1 item. A set of 5 possible item forms are given together with explanations and examples in Table \ref{tab:output-examples}. The item forms consist of comparisons over $m \in M$ or $p \in P$, labels $l \in L$, and whether an edge $e \in E$ exists for a pair of $v \in V$.


\section{Learning Semantic Association Rules - AE SemRL} \label{sec:ae-semrl}


\textbf{A}uto\textbf{E}ncoder-based \textbf{Sem}antic association \textbf{R}ule \textbf{L}earning (AE SemRL) is a method and an algorithm that facilitates Autoencoders to learn representations of semantically enriched time series data in the form of transactions and extract association rules from the representation. 

\subsection{Autoencoder Architecture}\label{sec:autoencoder-arch}


In this study, we utilized a \textit{denoising} under-complete Autoencoder~\citep{denoisingautoencoder}. It aims to reconstruct the input from its noisy variant, which is created by adding a noise value to the input. In this way, the model becomes more robust to noise in the input data. During the training process, $tanh(z) = \frac{e^z - e^{-z}}{e^z + e^{-z}}$ is preferred in the hidden layers and $softmax(z_i) = \frac{e^{z_i}}{\sum_{j=1}^{n} e^{z_j}}$ preferred at the output layer, as activation functions. The softmax function is applied per category of features so that for each category, probabilities per class values are obtained. As the lost function, aggregated binary cross entropy loss, $BCE\_Loss = \frac{1}{n} \sum_{i=1}^{n} -(y_i log(p_i) + (1-y_i) log(1-p_i))$, is applied to each feature to calculate the loss between Autoencoder reconstruction and noise-free input.

\subsection{Pipeline}

\begin{figure}[h]
    \centering
    \includegraphics[width=0.99\textwidth]{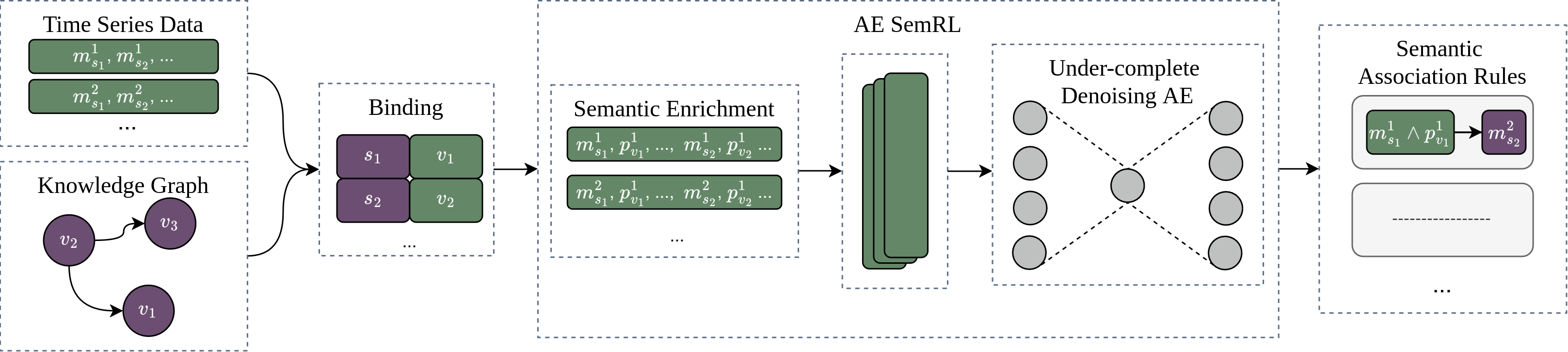}
    \caption{Full pipeline of operations for AE SemRL.}
    \label{fig:pipeline}
\end{figure}

Figure \ref{fig:pipeline} depicts the full pipeline of operations for AE SemRL. Time series data is discretized using equal frequency discretization method~\citep{foorthuis2020impact}, also to be comparable to baseline FP-Growth where we also use the same pre-discretization. Binding B is utilized to generate a semantically enriched set of transactions from time series data. Let $j$ be the number of time series data sources in $S$, $i$ be the number of semantic property values in $U$ mapped to each $s_{1..j}$, $z$ be the number of classes per input feature for simplicity, and $n$ be the number of transactions. In practice, $i$ and $z$ usually are different per $s_{1..j}$, and property values $p \in U$ can be different per transaction if G changes over time. Property values from neighbors of node $v$, i.e., $p_{v'_j}^k$ where $\exists e \in E:v \rightarrow v'$, can also be in the transaction set depending on the application. 

Semantically enriched transactions are then one-hot encoded and fed into the Autoencoder. AE SemRL keeps track of which data is fed into which neuron.

\noindent Input transactions to the Autoencoder look as follows: \\

$[\{m1_{s_1}^1, m1_{s_1}^2, ..., m1_{s_1}^z, ..., m1_{s_j}^z, p1_{{s_1}_1}^1, ..., p1_{{s_1}_1}^z, ..., p1_{{s_1}_i}^z, ..., p1_{{s_j}_i}^z\}, \\
\indent \{m2_{s_1}^1, m2_{s_1}^2, ..., m2_{s_1}^z, ..., m2_{s_j}^z, p2_{{s_1}_1}^1, ..., p2_{{s_1}_1}^z, ..., p2_{{s_1}_i}^z, ..., p2_{{s_j}_i}^z\} \\
\indent ... \\ 
\indent \{mn_{s_1}^1, mn_{s_1}^2, ..., mn_{s_1}^z, ..., mn_{s_j}^z, pn_{{s_1}_1}^1, ..., pn_{{s_1}_1}^z, ..., pn_{{s_1}_i}^z, ..., pn_{{s_j}_i}^z\}]$ \\


\subsection{Rule Extraction from Autoencoders} \label{sec:rule-extraction-autoencoder}

\textbf{Intuition:} AE SemRL exploits the reconstruction loss of a trained Autoencoder to learn associations. If reconstruction for an input vector with marked features is more successful than a certain threshold (\textit{similarity threshold}) then we say that the marked features imply the successfully reconstructed features. Marking features is done by assigning 1 (100\%) probability to a certain class value for a feature, 0 to the other classes for the same feature, and then assigning equal probabilities to the rest of the features in an input vector. 

\begin{figure}[h]
    \centering
    \includegraphics[width=0.99\textwidth]{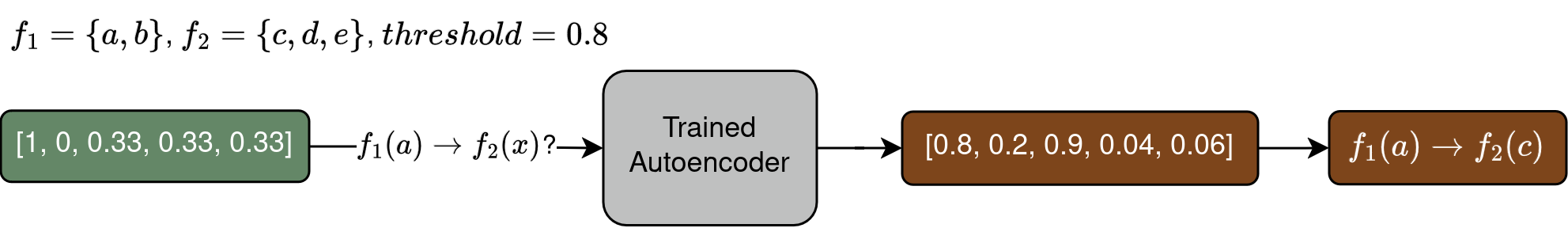}
    \caption{An illustration of association rule extraction from a trained Autoencoder.}
    \label{fig:example}
\end{figure}

\textbf{Example:} Figure \ref{fig:example} depicts an example rule extraction process. Assume that there are only two features in the input vector where one has 2 and the other has 3 possible class values, namely $f_1 = \{a,b\}$ and $f_2=\{c,d,e\}$. One-hot encoded version of $f_1$ and $f_2$ can be represented with 5 digits. Assume that we want to test whether $f_1(a) \rightarrow f_2(c)$, meaning whether $f_1$ being `$a$' implies $f_2$ being `$c$' or not, based on the similarity threshold of 0.8. In this case, we do a forward run on the trained Autoencoder with the input vector $[1,0,0.33,0.33,0.33]$. We call this a $test\_vector$. Assume that the output is $[0.8,0.2,0.9,0.04,0,06]$. The third output digit that corresponds to $f_2(c)$ is bigger than the threshold, 0.9. Therefore, we say that $f_1(a) \rightarrow f_2(c)$.

\textbf{Algorithm:} The algorithm for extracting association rules from a trained Autoencoder is given in Algorithm \ref{alg:ae-semrl}. The parameters are a trained Autoencoder (\textit{ae\_model}), a similarity threshold (\textit{similarity\_threshold}), and a maximum number of antecedents (\textit{max\_antecedents}) that the rules will contain. Based on the $max\_antecedents$, in line 3, the algorithm creates combinations of features to be tested. Instead of marking each combination of input features to create test vectors, it goes through all unique input vectors (line 4) to create test vectors, since a pattern that is not in the input vectors can not be an association. In lines 6 and 7, it creates a test vector (as demonstrated in the previous examples), where candidate antecedents are marked features, and the rest of the features have equal probabilities. In lines 8-12, it performs a forward run on the \textit{ae\_model} with the test vector and compares the probabilities of class values per feature in the Autoencoder output with the similarity threshold. If the probability is bigger than the threshold, then it constructs the rule as demonstrated in the examples. In line 13, it returns the extracted association rules.

\begin{algorithm}[h]
	\caption{AE SemRL} 
    \begin{algorithmic}[1]
    \Procedure{AESemRL}{ae\_model, similarity\_threshold, max\_antecedent}
        \State rules = []
        \State candidate\_antecedents = combinations(ae\_model.features, max\_antecedents)
        \For {input\_vector in unique(ae\_model.input\_vectors)}
            \For {feature\_list in candidate\_antecedent}
                \State test\_vector = init\_vector\_with\_equal\_probabilities(ae\_model.features)
                \State test\_vector[feature\_list] = input\_vector[feature\_list]
                \If {test\_vector not tested before}:
                    \State output = ae\_model.forward(test\_vector)
                    \For {feature in output}:
                        \If {feature.class\_with\_max\_prob() $>$ similarity\_threshold}:
                            \State rules.append(\{antecedent: feature\_list, consequent: feature\})
                        \EndIf
                    \EndFor
                \EndIf
            \EndFor
        \EndFor
        \State \Return rules
    \EndProcedure
    \end{algorithmic}
    \label{alg:ae-semrl}
\end{algorithm}

\section{Evaluation} \label{sec:experiments}

This section includes an evaluation of the AE SemRL on 3 different open-source datasets and 2 baselines. The entire source code is written in Python and is available online (anonymously) including the source codes for knowledge graph construction, parsing time series data, baselines, and AE SemRL\footnote{https://anonymous.4open.science/r/semantic-association-rule-learning-5372}.

\subsection{Setup}

3 open-source time series datasets with associated semantics from two different domains, water networks, and energy, are used in the evaluation. LeakDB~\citep{leakdb} is an artificially generated realistic dataset for leak detection in water distribution networks. It contains time series data from 96 sensors of various types, and semantic information such as the formation of the network, sensor placement, and properties of network components, i.e., pipes. L-Town\citep{ltown} is another dataset in the water distribution networks domain with the same characteristics. It has 118 sensors. The third is LBNL Fault Detection and Diagnostics Datasets~\citep{lbnl} that contains time series data from 29 sensors and semantics for Heating, Ventilation, and Air Conditioning (HVAC) systems such as the placement of the sensors and structure of the HVAC system. A knowledge graph is created per dataset.

An Autoencoder with the given architecture (Section \ref{sec:autoencoder-arch}) is trained for each dataset. During training, the following parameters were found to be performing best based on a grid search: learning rate is set to $5e^{-3}$, the models are trained for 20 epochs, Adam~\citep{kingma2014adam} optimizer is used for gradient optimization with a weight decay of $2e^{-8}$, and the noise factor for the denoising Autoencoder is $0.5$.

\subsection{Evaluation Metrics}

The metrics that are used in the evaluation are \textit{number of rules}, \textit{execution time} for learning rules (excluding KG construction, semantic enrichment, model training, etc.), \textit{rule overlap} and \textit{rule quality}. Measuring rule quality is the most common way to evaluate ARM algorithms~\citep{kaushik2023numerical,telikani2020survey}. We selected the following most common rule quality criteria that are used to measure the implied strength of association in a rule (lift, leverage, Zhang's metric) since our hypothesis is to show that Autoencoders can learn associations among its input features: 

\begin{itemize}
        \item \textbf{Support}: Percentage of transactions with a certain item or rule, among all transactions (D): $support(X \rightarrow Y) = \frac{|X \cup Y|}{|D|}$
        \item \textbf{Confidence}: Conditional probability of a rule, e.g., given the transactions with the antecedent X in, the probability of having the consequent Y in the same transaction set: $confidence(X \rightarrow Y) = \frac{|X \cup Y|}{|X|}$
        \item \textbf{Lift}: Frequency of a rule $X \rightarrow Y$ occurring compared to the case where they were statistically independent. A score of 1 indicates independence and $>1$ indicates association: $lift(X \rightarrow Y) = \frac{confidence(X \rightarrow Y)}{support(Y)}$
        \item \textbf{Leverage}: Difference between antecedent and consequent appearing together and the expected frequency of their independence. A score of $0$ indicates independence while a score of $>0$ indicates association. $leverage(X \rightarrow Y) = support(X \rightarrow Y) - (support(X) \times support(Y))$
        \item \textbf{Zhang's Metric~\citep{zhangsmetric}}: This metric also considers the case in which the consequent appears alone in the transaction set, besides their co-occurrence, and therefore measures dissociation as well. A score of $>0$ indicates an association, 0 indicates independence and $<0$ indicates dissociation: $zm(X \rightarrow Y) = \frac{confidence(X \rightarrow Y) - confidence(X' \rightarrow C)}{max(confidence(X \rightarrow Y), confidence(X' \rightarrow Y))}$ in which $X'$ refers to the absent of $X$ in the transaction set.
\end{itemize}

\subsection{Baselines} \label{sec:baselines}

Two baselines are used for evaluation. The first one is the only existing approach for learning semantic association rules from time series data~\citep{karabulut2023semantic} that is based on the well-known FP-Growth~\citep{han2000mining} algorithm. It relies on a pre-discretization of the time series data for mining, and it is implemented using MLxtend Python package~\citep{raschkas_2018_mlxtend}. FP-Growth takes a minimum support and confidence threshold value as parameters and produces rules that have higher values than the thresholds. The second baseline is a state-of-the-art optimization-based ARM method, based on highly influential Harris' Hawk Optimization (HHO)~\citep{heidari2019harris} algorithm, implemented using NiaPy and NiaARM Python packages~\citep{vrbanvcivc2018niapy,stupan2022niaarm}. HHO does not require a pre-discretization step, takes an initial population size (random set of solutions, in this case, association rules), and several iterations for optimization steps, and produces a set of association rules optimizing a fitness function (rules with high confidence is mined with HHO). 

\subsection{Experiments}

This section includes a set of experiments to evaluate and compare AE SemRL with baselines FP-Growth and HHO from different aspects.


\subsubsection{How fast is AE SemRL in comparison to baselines?} \label{sec:runtime-experiment}

The time complexity of FP-Growth is $D(2N+D+2^D)+2^D$~\citep{yuan2012research}, where $D$ refers to the data dimension and $N$ refers to the number of transactions, with the most expensive operation being finding maximum frequent sets in the data, $O(2^D)$. The time complexity of HHO is $O(N×(T + TD + 1))$~\citep{heidari2019harris} with T being the number of iterations. AE SemRL, on the other hand, has the time complexity of $O(N^A)$ where $A$ is the number of antecedents. However, the time complexity depends highly on the algorithm parameters such as support and confidence for FP-Growth, initial population size for HHO, maximum number of antecedents for AE SemRL, as well as dataset characteristics such as the number of unique items. Therefore, we provide an execution time analysis below.

\begin{figure}[b!]
    \centering
    \includegraphics[width=\textwidth]{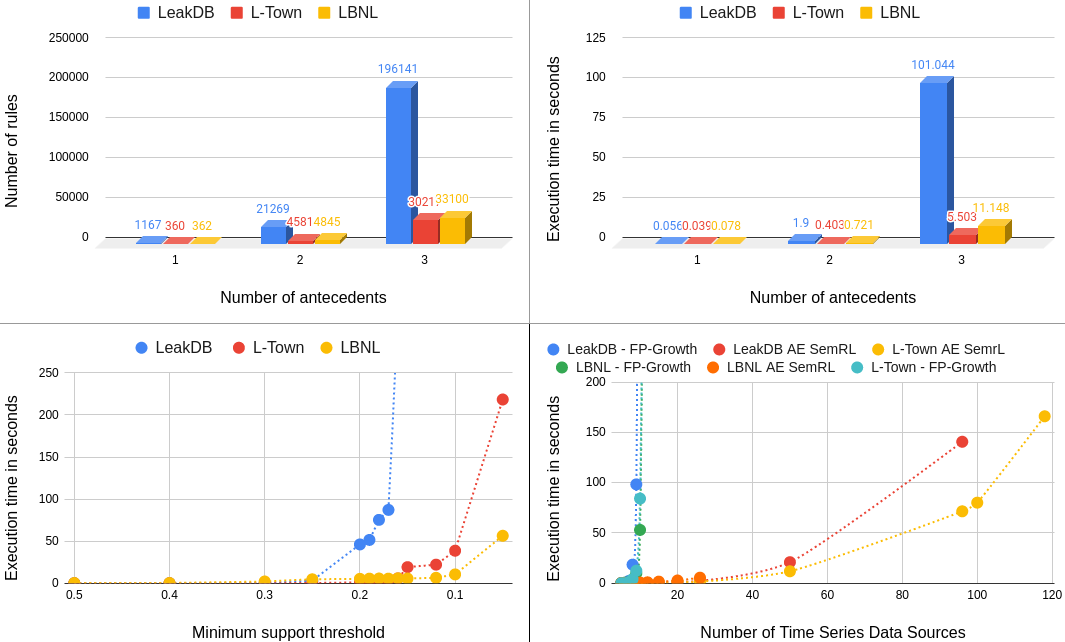}
    \caption{(Bottom-Left) Execution time of FP-Growth (confidence = 0.8). (Bottom-Right) Effect of the number of time series data sources on execution time (AE SemRL's similarity threshold = 0.5, support=0.25, confidence=0.8 for FP-Growth). (Top-Left) The number of rules learned by AE SemRL. (Top-Right) Execution time of AE SemRL per different number of antecedents.}
    \label{fig:fpgrowth-vs-aesemrl-runtime}
\end{figure}

First, we show that it is impractical to run FP-Growth on the full datasets. The hardware used in the experiment is a Lenovo ThinkPad X1 Yoga Gen 7 personal computer with 12th Gen Intel® Core™ i5-1240P × 16 model CPU and 16,0 GiB memory. The bottom-right graph in Figure \ref{fig:fpgrowth-vs-aesemrl-runtime} shows the results of an execution time experiment of both approaches on all 3 datasets for learning semantic association rules with 1 antecedent. This experiment is carried out by randomly picking time series data sources 10 times, and calculating average execution times for each of the points on the graph. The results show that FP-Growth is highly sensitive to the width of each transaction and the execution time of the algorithm becomes impractical after considering roughly 10 time series data sources and associated semantics. On the other hand, AE SemRL shows a more stable behavior and is resilient to the choice of data source selection. The graph on the bottom-left side of Figure \ref{fig:fpgrowth-vs-aesemrl-runtime} shows the effect of the minimum support threshold on FP-Growth's execution time. Depending on the dataset, after reducing the threshold (to somewhere between 0.2 and 0.01), FP-Growth starts to perform poorly in terms of execution time. 

Top-left and top-right of Figure \ref{fig:fpgrowth-vs-aesemrl-runtime} shows the effect of the number of antecedents parameter in AE SemRL on the execution time and the number of rules learned. Due to the combination operation on the number of antecedents to create test vectors in the AE SemRL algorithm (line 3 in Algorithm \ref{alg:ae-semrl}), we see a factorial-grade increase in the execution time. Overall, these experiments show that FP-Growth is sensitive to the minimum support threshold and the number of time series data sources, while AE SemRL is sensitive to the number of antecedent parameters in terms of their execution time. 

Since the execution time of HHO is directly influenced by the manually set maximum iterations parameter, we base our evaluation on rule quality. We first run AE SemRL, measure rule quality, and then run HHO for approximately enough time to get similar confidence values to AE SemRL. The initial population of HHO is set to the number of rules obtained by the AE SemRL approach. Table \ref{table:hho-vs-aesemrl} shows the result of this analysis. The results show that HHO takes 100s of times longer to compute and produces 10s of times more rules with lower implied association strength on average.

\begin{table}[htp]
    \centering
    \caption{Comparison of AE SemRL and HHO (p=initial population, i=max iteration) on the 3 datasets.}
    \label{table:hho-vs-aesemrl}
    \begin{tabular}{p{0.07\textwidth}p{0.2\textwidth}p{0.07\textwidth}p{0.09\textwidth}p{0.05\textwidth}p{0.07\textwidth}p{0.07\textwidth}p{0.07\textwidth}p{0.07\textwidth}}
    \toprule
        Dataset & Approach & Support & Confidence & Lift & Leverage & Zhang's Metric &  Exec. time & Rule Count \\ 
        \midrule
        \multirow{2}{4em}{LeakDB} & AE SemRL & \textbf{0.263} & \textbf{0.903} & 1.467 & \textbf{0.015} & \textbf{0.41} & \textbf{0.056} & 450 \\
        & HHO(p=450, i=20000) & 0.208 & 0.827 & \textbf{1.838} & 0.014 & 0.181 & 38.915 & \textbf{8457} \\ \hline & \\[-2ex]
        \multirow{2}{4em}{L-Town} & AE SemRL & \textbf{0.333} & \textbf{0.877} & 0.157 & \textbf{0.008} & \textbf{0.443} & \textbf{0.039} & 200 \\
        & HHO(p=200, i=10000) & 0.285 & 0.849 & \textbf{1.629} & 0.002 & 0.04 & 14.908 & \textbf{3975} \\ \hline & \\[-2ex]
        \multirow{2}{4em}{LBNL} & AE SemRL & \textbf{0.356} & \textbf{0.973} & 1.024 & 0.0006 & \textbf{0.374} & \textbf{0.078} & 206 \\
        & HHO(p=206, i=50000) & \textbf{0.417} & 0.817 & \textbf{1.54} & \textbf{0.0021} & 0.099 & 180.551 & \textbf{21851} \\
        \bottomrule
    \end{tabular}
\end{table}

\subsubsection{How good are the rules found by AE SemRL?}

As FP-Growth is based on counting items in the transaction dataset, we assume that it can already find strong associations based on a given criterion. The goal of AE SemRL is to approximate FP-Growth as much as possible. To avoid the unfeasible runtime of FP-Growth (see previous experiment), we only picked 7 time series data sources randomly for this experiment. It is repeated 10 times to prevent bias in the choice of data source selection and average results are presented. FP-Growth is run with different minimum support thresholds as this value has a direct impact on the rule quality, and the AE SemRL is run with a 0.5 similarity threshold. 

Table \ref{tab:rule-quality-comparison} shows that the rules learned by AE SemRL have high confidence (87\%+).  Lift, leverage and Zhang's metrics for AE SemRL show that the rules imply association, and not only statistical significance. The quality of the rules mined by FP-Growth with relatively higher support values, is significantly higher than the AE SemRL but produces a low number of rules. When run with lower support values, this difference becomes smaller and the number of rules increases. According to the rule overlap results, a significant amount of association rules (50\%+ on average) mined by FP-Growth is also learned by AE SemRL. 

A rule quality comparison of AE SemRL and HHO is already given in Table \ref{table:hho-vs-aesemrl}. Note that the rule overlap criterion does not apply to HHO, since it does not do a pre-discretization, and hence intervals for numerical properties are different. An additional experiment showing how similarity threshold effects rule quality in AE SemRL is given in Appendix \ref{appendix:additional-experiments}.

\begin{table}[htp]
    \centering
    \caption{Comparison of AE SemRL and FP-Growth (FPG, sup=minimum support threshold) on the 3 datasets.}
    \label{tab:rule-quality-comparison}
    \begin{tabular}{p{0.07\textwidth}p{0.13\textwidth}p{0.07\textwidth}p{0.09\textwidth}p{0.07\textwidth}p{0.07\textwidth}p{0.07\textwidth}p{0.07\textwidth}p{0.07\textwidth}}
        \toprule
        Dataset & Approach & Support & Confidence & Lift & Leverage & Zhang's Metric & Rule Count & Rule Overlap\\ 
        \midrule
        \multirow{4}{5em}{LeakDB} & AE SemRL & 0.263 & 0.903 & 1.467 & 0.015 & 0.41  & \textbf{450.1} & - \\
        & FPG(sup$\geq$0.4) & \textbf{0.457} & 1 & 2.131 &	0.197 &	0.93 & 5.2 & \textbf{77\%} \\
        & FPG(sup$\geq$0.3) & 0.334 & 0.993 & 2.976 & \textbf{0.219} & \textbf{0.99} & 40.8 & 41\% \\ 
        & FPG(sup$\geq$0.2) & 0.258 & \textbf{0.99} & \textbf{3.882} & 0.185 & \textbf{0.99} & 77.4 & 31\% \\ \hline & \\[-2ex]
        \multirow{4}{5em}{L-Town} & AE SemRL & 0.333 & 0.877 & 1.157 & 0.008 & 0.443 & \textbf{199.6} & -  \\
        & FPG(sup$\geq$0.4)& \textbf{0.701} & \textbf{0.985} & 1.528 & \textbf{0.11} & \textbf{0.95} & 16.2 & 68\% \\
        & FPG(sup$\geq$0.3)& 0.545 & 0.968 & 1.566 & 0.089 & 0.785 & 28 & 65\% \\ 
        & FPG(sup$\geq$0.2)& 0.366 & 0.97 & \textbf{1.67} & 0.079 & 0.611 & 78 & \textbf{82\%} \\ \hline & \\[-2ex]
        \multirow{4}{5em}{LBNL} & AE SemRL & 0.356 & 0.973 & 1.024 & 0.0006 & 0.374 & \textbf{206.5} & - \\
        & FPG(sup$\geq$0.4)& \textbf{0.904} & 0.989 & 1.029 & 0.012 & \textbf{0.932} & 27.6 & \textbf{62\%} \\
        & FPG(sup$\geq$0.3)& 0.843 & \textbf{0.997} & 1.078 & 0.018 & 0.885 & 35.4 & 60\% \\ 
        & FPG(sup$\geq$0.2)& 0.587 & 0.99 & \textbf{1.926} & \textbf{0.086} & 0.875 & 47.6 & 55\% \\
        \bottomrule
    \end{tabular}
\end{table}

\section{Discussion} \label{sec:discussion}

This section discusses the experiment results and future research directions.


\textbf{Need for learning-based approaches.} State-of-the-art ARM approaches perform poorly on big high-dimensional data. Both FP-Growth and HHO are highly sensitive to the number of time series data sources as shown in the experiments. Furthermore, FP-Growth is highly sensitive to lower minimum support thresholds which allows the algorithm to capture rare but consistent events. HHO, on the other hand, has to run a long time compared to AE SemRL to be able to extract high-quality association rules. Therefore, there is a great research opportunity on the learning-based approaches to replace ARM algorithms.

\textbf{Learning associations with Autoencoders.} The experiments showed that Autoencoders are capable of learning associations in the input data. However, AE SemRL has high execution times when run to learn rules with a relatively higher number of antecedents. Rules with a bigger set of antecedents refer to more specific cases in the data compared to the rules with a smaller set of antecedents. Therefore, we can say that AE SemRL can learn more generalizable patterns faster than more specific patterns in the data. A natural next step is to experiment with different levels of semantic properties in the antecedent of a rule to decide how much semantics is enough to achieve a certain task.

\textbf{Extracting association rules from latent representations.} Autoencoders are one way to encode associations in the data. However, different representation learning methods should also be tested for the task of learning associations. For instance, graph neural networks are better at capturing a graph's structure and therefore are a good candidate to learn associations from a knowledge graph. Another aspect is to extract associations from a representation. AE SemRL offers a straightforward intuitive method, however, more sophisticated methods, e.g., based on output probability distributions are to be investigated.

\textbf{Post-processing of the semantic association rules.} This paper, and most articles on ARM, offers an evaluation based on the quality of the rules. Another way to evaluate the rules is to test them for a certain task such as leak detection for water networks, fault detection for energy use cases, or anomaly detection in general. This requires building hypotheses using the rules and testing them on unseen data. Another open research question is to understand the strengths and weaknesses of semantic association rules on such tasks.

\section{Conclusion} \label{sec:conclusion}

This study showed that Autoencoders can learn association rules in a semantically enriched time series dataset. Rules in this form are called semantic association rules which are more generically applicable and explainable. We proposed a pipeline of operations to semantic association rule learning and an algorithm named AE SemRL that can extract associations from an Autoencoder. Our evaluation results based on 3 datasets from 2 domains show that the quality of the rules learned by AE SemRL implies associations and the algorithm can run much faster than the state-of-the-art ARM algorithms in many scenarios.

These findings led to new research questions including investigating learning associations with other neural network architectures and finding other ways to extract associations from latent representations that are more efficient or can lead to stronger associations. Once learned, semantic association rules can be used in many applications such as leak detection in water networks, and fault diagnosis in the energy domain. Another research direction is to post-process the rules to find a subset that can be used for a certain task or to adopt the learning process (e.g., via a feedback mechanism) for a certain task.

\textbf{Acknowledgements.} This work has received support from The Dutch Research Council (NWO), in the scope of the Digital Twin for Evolutionary Changes in water networks (DiTEC) project, file number 19454.

%
%
%

\bibliographystyle{unsrtnat}
\bibliography{main}

\begin{thebibliography}{34}
\providecommand{\natexlab}[1]{#1}
\providecommand{\url}[1]{\texttt{#1}}
\expandafter\ifx\csname urlstyle\endcsname\relax
  \providecommand{\doi}[1]{doi: #1}\else
  \providecommand{\doi}{doi: \begingroup \urlstyle{rm}\Url}\fi

\bibitem[Nassif et~al.(2021)Nassif, Talib, Nasir, and Dakalbab]{nassif2021machine}
Ali~Bou Nassif, Manar~Abu Talib, Qassim Nasir, and Fatima~Mohamad Dakalbab.
\newblock Machine learning for anomaly detection: A systematic review.
\newblock \emph{Ieee Access}, 9:\penalty0 78658--78700, 2021.

\bibitem[Khanal et~al.(2020)Khanal, Prasad, Alsadoon, and Maag]{khanal2020systematic}
Shristi~Shakya Khanal, PWC Prasad, Abeer Alsadoon, and Angelika Maag.
\newblock A systematic review: machine learning based recommendation systems for e-learning.
\newblock \emph{Education and Information Technologies}, 25:\penalty0 2635--2664, 2020.

\bibitem[Jiang et~al.(2018)Jiang, Kwong, Park, and Yu]{jiang2018multi}
Huimin Jiang, Chun~Kit Kwong, WY~Park, and Kai~Ming Yu.
\newblock A multi-objective pso approach of mining association rules for affective design based on online customer reviews.
\newblock \emph{Journal of Engineering Design}, 29\penalty0 (7):\penalty0 381--403, 2018.

\bibitem[Agrawal et~al.(1994)Agrawal, Srikant, et~al.]{agrawal1994fast}
Rakesh Agrawal, Ramakrishnan Srikant, et~al.
\newblock Fast algorithms for mining association rules.
\newblock In \emph{Proc. 20th int. conf. very large data bases, VLDB}, volume 1215, pages 487--499. Santiago, Chile, 1994.

\bibitem[Han et~al.(2000)Han, Pei, and Yin]{han2000mining}
Jiawei Han, Jian Pei, and Yiwen Yin.
\newblock Mining frequent patterns without candidate generation.
\newblock \emph{ACM sigmod record}, 29\penalty0 (2):\penalty0 1--12, 2000.

\bibitem[Kaushik et~al.(2023)Kaushik, Sharma, Fister~Jr, and Draheim]{kaushik2023numerical}
Minakshi Kaushik, Rahul Sharma, Iztok Fister~Jr, and Dirk Draheim.
\newblock Numerical association rule mining: A systematic literature review.
\newblock \emph{arXiv preprint arXiv:2307.00662}, 2023.

\bibitem[Telikani et~al.(2020)Telikani, Gandomi, and Shahbahrami]{telikani2020survey}
Akbar Telikani, Amir~H Gandomi, and Asadollah Shahbahrami.
\newblock A survey of evolutionary computation for association rule mining.
\newblock \emph{Information Sciences}, 524:\penalty0 318--352, 2020.

\bibitem[Sarker(2021)]{sarker2021machine}
Iqbal~H Sarker.
\newblock Machine learning: Algorithms, real-world applications and research directions.
\newblock \emph{SN computer science}, 2\penalty0 (3):\penalty0 160, 2021.

\bibitem[Qu et~al.(2020)Qu, Chen, Xhonneux, Bengio, and Tang]{qu2020rnnlogic}
Meng Qu, Junkun Chen, Louis-Pascal Xhonneux, Yoshua Bengio, and Jian Tang.
\newblock Rnnlogic: Learning logic rules for reasoning on knowledge graphs.
\newblock \emph{arXiv preprint arXiv:2010.04029}, 2020.

\bibitem[Hogan et~al.(2021)Hogan, Blomqvist, Cochez, d'Amato, de~Melo, Guti\'errez, Kirrane, Labra~Gayo, Navigli, Neumaier, Ngonga~Ngomo, Polleres, Rashid, Rula, Schmelzeisen, Sequeda, Staab, and Zimmermann]{kg-book}
Aidan Hogan, Eva Blomqvist, Michael Cochez, Claudia d'Amato, Gerard de~Melo, Claudio Guti\'errez, Sabrina Kirrane, Jos\'e~Emilio Labra~Gayo, Roberto Navigli, Sebastian Neumaier, Axel-Cyrille Ngonga~Ngomo, Axel Polleres, Sabbir~M. Rashid, Anisa Rula, Lukas Schmelzeisen, Juan~F. Sequeda, Steffen Staab, and Antoine Zimmermann.
\newblock \emph{Knowledge Graphs}.
\newblock Number~22 in Synthesis Lectures on Data, Semantics, and Knowledge. Springer, 2021.
\newblock ISBN 9783031007903.
\newblock \doi{10.2200/S01125ED1V01Y202109DSK022}.
\newblock URL \url{https://kgbook.org/}.

\bibitem[Karabulut et~al.(2023{\natexlab{a}})Karabulut, Pileggi, Groth, and Degeler]{karabulut2023ontologies}
Erkan Karabulut, Salvatore~F Pileggi, Paul Groth, and Victoria Degeler.
\newblock Ontologies in digital twins: A systematic literature review.
\newblock \emph{Future Generation Computer Systems}, 2023{\natexlab{a}}.

\bibitem[Greco et~al.(2021)Greco, Ritrovato, and Vento]{greco2021use}
Luca Greco, Pierluigi Ritrovato, and Mario Vento.
\newblock On the use of semantic technologies for video analytics.
\newblock \emph{Journal of Ambient Intelligence and Humanized Computing}, 12:\penalty0 567--587, 2021.

\bibitem[Berteloot et~al.(2023)Berteloot, Khoury, and Durand]{berteloot2023association}
Th{\'e}ophile Berteloot, Richard Khoury, and Audrey Durand.
\newblock Association rules mining with auto-encoders.
\newblock \emph{arXiv preprint arXiv:2304.13717}, 2023.

\bibitem[Kishore et~al.(2021)Kishore, Bhushan, and Suneetha]{kishore2021applications}
Sai Kishore, Vikram Bhushan, and KR~Suneetha.
\newblock Applications of association rule mining algorithms in deep learning.
\newblock In \emph{Computer Networks and Inventive Communication Technologies: Proceedings of Third ICCNCT 2020}, pages 351--362. Springer, 2021.

\bibitem[Barati et~al.(2017)Barati, Bai, and Liu]{barati2017mining}
Molood Barati, Quan Bai, and Qing Liu.
\newblock Mining semantic association rules from rdf data.
\newblock \emph{Knowledge-Based Systems}, 133:\penalty0 183--196, 2017.

\bibitem[d’Amato(2020)]{d2020machine}
Claudia d’Amato.
\newblock Machine learning for the semantic web: Lessons learnt and next research directions.
\newblock \emph{Semantic Web}, 11\penalty0 (1):\penalty0 195--203, 2020.

\bibitem[Karabulut et~al.(2023{\natexlab{b}})Karabulut, Degeler, and Groth]{karabulut2023semantic}
Erkan Karabulut, Victoria Degeler, and Paul Groth.
\newblock Semantic association rule learning from time series data and knowledge graphs.
\newblock In \emph{Proceedings of the 2nd International Workshop on Semantic Industrial Information Modelling (SemIIM 2023) co-located with 22nd International Semantic Web Conference (ISWC 2023)}, pages 1--7, 2023{\natexlab{b}}.

\bibitem[Simsek et~al.(2021)Simsek, Yildirim~Okay, and Ozdemir]{simsek2021deep}
Mehmet~Ulvi Simsek, Feyza Yildirim~Okay, and Suat Ozdemir.
\newblock A deep learning-based cep rule extraction framework for iot data.
\newblock \emph{The Journal of Supercomputing}, 77:\penalty0 8563--8592, 2021.

\bibitem[Chakraborty et~al.(2020)Chakraborty, Biswas, and Purkayastha]{chakraborty2020rule}
Manomita Chakraborty, Saroj~Kumar Biswas, and Biswajit Purkayastha.
\newblock Rule extraction from neural network trained using deep belief network and back propagation.
\newblock \emph{Knowledge and Information Systems}, 62:\penalty0 3753--3781, 2020.

\bibitem[Patel et~al.(2022)]{patel2022innovative}
Harvendra~Kumar Patel et~al.
\newblock An innovative approach for association rule mining in grocery dataset based on non-negative matrix factorization and autoencoder.
\newblock \emph{Journal of Algebraic Statistics}, 13\penalty0 (3):\penalty0 2898--2905, 2022.

\bibitem[Chen and Guo(2023)]{chen2023auto}
Shuangshuang Chen and Wei Guo.
\newblock Auto-encoders in deep learning—a review with new perspectives.
\newblock \emph{Mathematics}, 11\penalty0 (8):\penalty0 1777, 2023.

\bibitem[Tama{\v{s}}auskait{\.e} and Groth(2023)]{tamavsauskaite2023defining}
Gyt{\.e} Tama{\v{s}}auskait{\.e} and Paul Groth.
\newblock Defining a knowledge graph development process through a systematic review.
\newblock \emph{ACM Transactions on Software Engineering and Methodology}, 32\penalty0 (1):\penalty0 1--40, 2023.

\bibitem[Vincent et~al.(2008)Vincent, Larochelle, Bengio, and Manzagol]{denoisingautoencoder}
Pascal Vincent, Hugo Larochelle, Yoshua Bengio, and Pierre-Antoine Manzagol.
\newblock Extracting and composing robust features with denoising autoencoders.
\newblock In \emph{Proceedings of the 25th international conference on Machine learning}, pages 1096--1103, 2008.

\bibitem[Foorthuis(2020)]{foorthuis2020impact}
Ralph Foorthuis.
\newblock The impact of discretization method on the detection of six types of anomalies in datasets.
\newblock \emph{arXiv preprint arXiv:2008.12330}, 2020.

\bibitem[Vrachimis et~al.(2018)Vrachimis, Kyriakou, et~al.]{leakdb}
Stelios~G Vrachimis, Marios~S Kyriakou, et~al.
\newblock Leakdb: a benchmark dataset for leakage diagnosis in water distribution networks:(146).
\newblock In \emph{WDSA/CCWI Joint Conference Proceedings}, volume~1, 2018.

\bibitem[Vrachimis et~al.(2020)Vrachimis, Eliades, Taormina, Ostfeld, Kapelan, Liu, Kyriakou, Pavlou, Qiu, and Polycarpou]{ltown}
SG~Vrachimis, DG~Eliades, R~Taormina, A~Ostfeld, Z~Kapelan, S~Liu, MS~Kyriakou, P~Pavlou, M~Qiu, and M~Polycarpou.
\newblock Dataset of battledim: Battle of the leakage detection and isolation methods.
\newblock In \emph{Proc., 2nd Int CCWI/WDSA Joint Conf. Kingston, ON, Canada: Queen's Univ}, 2020.

\bibitem[Granderson et~al.(2022)Granderson, Lin, Chen, Casillas, Im, Jung, Benne, Ling, Gorthala, Wen, Chen, Huang, , and Vrabie]{lbnl}
Jessica Granderson, Guanjing Lin, Yimin Chen, Armando Casillas, Piljae Im, Sungkyun Jung, Kyle Benne, Jiazhen Ling, Ravi Gorthala, Jin Wen, Zhelun Chen, Sen Huang, , and Draguna. Vrabie.
\newblock Lbnl fault detection and diagnostics datasets, 08 2022.
\newblock URL \url{https://data.openei.org/submissions/5763}.

\bibitem[Kingma and Ba(2014)]{kingma2014adam}
Diederik~P Kingma and Jimmy Ba.
\newblock Adam: A method for stochastic optimization.
\newblock \emph{arXiv preprint arXiv:1412.6980}, 2014.

\bibitem[Yan et~al.(2009)Yan, Zhang, and Zhang]{zhangsmetric}
Xiaowei Yan, Chengqi Zhang, and Shichao Zhang.
\newblock Confidence metrics for association rule mining.
\newblock \emph{Applied Artificial Intelligence}, 23\penalty0 (8):\penalty0 713--737, 2009.

\bibitem[Raschka(2018)]{raschkas_2018_mlxtend}
Sebastian Raschka.
\newblock Mlxtend: Providing machine learning and data science utilities and extensions to python’s scientific computing stack.
\newblock \emph{The Journal of Open Source Software}, 3\penalty0 (24), April 2018.
\newblock \doi{10.21105/joss.00638}.
\newblock URL \url{https://joss.theoj.org/papers/10.21105/joss.00638}.

\bibitem[Heidari et~al.(2019)Heidari, Mirjalili, Faris, Aljarah, Mafarja, and Chen]{heidari2019harris}
Ali~Asghar Heidari, Seyedali Mirjalili, Hossam Faris, Ibrahim Aljarah, Majdi Mafarja, and Huiling Chen.
\newblock Harris hawks optimization: Algorithm and applications.
\newblock \emph{Future generation computer systems}, 97:\penalty0 849--872, 2019.

\bibitem[Vrban{\v{c}}i{\v{c}} et~al.(2018)Vrban{\v{c}}i{\v{c}}, Brezo{\v{c}}nik, Mlakar, Fister, and Fister]{vrbanvcivc2018niapy}
Grega Vrban{\v{c}}i{\v{c}}, Lucija Brezo{\v{c}}nik, Uro{\v{s}} Mlakar, Du{\v{s}}an Fister, and Iztok Fister.
\newblock Niapy: Python microframework for building nature-inspired algorithms.
\newblock \emph{Journal of Open Source Software}, 3\penalty0 (23):\penalty0 613, 2018.

\bibitem[Stupan and Fister(2022)]{stupan2022niaarm}
{\v{Z}}iga Stupan and Iztok Fister.
\newblock Niaarm: a minimalistic framework for numerical association rule mining.
\newblock \emph{Journal of Open Source Software}, 7\penalty0 (77):\penalty0 4448, 2022.

\bibitem[Yuan and Ding(2012)]{yuan2012research}
Jingbo Yuan and Shunli Ding.
\newblock Research and improvement on association rule algorithm based on fp-growth.
\newblock In \emph{Web Information Systems and Mining: International Conference, WISM 2012, Chengdu, China, October 26-28, 2012. Proceedings}, pages 306--313. Springer, 2012.

\end{thebibliography}

\appendix

\section{Additional experiment} \label{appendix:additional-experiments}

\subsection{What is the effect of similarity threshold on the quality of the rules learned by AE SemRL?}

In this experiment, we investigate the effect of the similarity threshold of AE SemRL on the quality of the rules on the 3 datasets. The experiments are repeated 10 times on a subset of 7 randomly selected sensors each time (to avoid the impractical run time of Naive SemRL as shown in Section \ref{sec:runtime-experiment}), and the average results are presented. Table \ref{tab:similarity-threshold} shows the results with similarity threshold values varying between 0.9 and 0.5. The results show that when run with relatively lower similarity thresholds, rules learned by AE SemRL imply stronger associations based on the lift, leverage, and Zhang's metric values. Lower threshold values result in less statistical significance (lower support and confidence). 

\begin{table}[h]
    \centering
    \caption{Effect of similarity threshold on the rule quality on the 3 datasets.}
    \label{tab:similarity-threshold}
    \begin{tabular}{p{0.07\textwidth}p{0.09\textwidth}p{0.07\textwidth}p{0.09\textwidth}p{0.07\textwidth}p{0.07\textwidth}p{0.07\textwidth}}
        \toprule
        Dataset & Similarity Threshold & Support & Confidence & Lift & Leverage & Zhang's Metric \\ 
        \midrule
        \multirow{5}{5em}{LeakDB} & 0.9 & \textbf{0.277} & \textbf{0.951} & 1.49 & 0.017 & 0.427 \\
        & 0.8 & 0.267 & 0.919 & 1.59 & 0.019 & 0.458 \\
        & 0.7 & 0.26 & 0.896 & 1.657 & 0.021 & 0.47 \\
        & 0.6 & 0.253 & 0.875 & 1.735 & 0.022 & 0.49 \\ 
        \hline & \\[-2ex]
        
        \multirow{5}{5em}{L-Town} & 0.9 & \textbf{0.468} & \textbf{0.994} & 1.064 & 0.008 & 0.502 \\
        & 0.8 & 0.452 & 0.871 & 1.138 & 0.016 & 0.53 \\
        & 0.7 & 0.436 & 0.918 & 1.169 & 0.018 & 0.567 \\ 
        & 0.6 & 0.409 & 0.864 & 1.176 & 0.018 & 0.549 \\
         \hline & \\[-2ex]
         
         \multirow{5}{5em}{LBNL} & 0.9 & \textbf{0.42} & \textbf{0.985} & 1.044 & 0.004 & 0.447 \\ 
         & 0.8 & 0.416 & 0.976 & \textbf{1.057} & \textbf{0.005} & 0.449 \\
         & 0.7 & 0.416 & 0.967 & 1.05 & \textbf{0.005} & 0.446 \\
         & 0.6 & 0.416 & 0.96 & 1.05 & \textbf{0.005} & \textbf{0.454} \\
         & 0.5 & 0.413 & 0.955 & 1.052 & \textbf{0.005} & 0.453 \\ 
         \bottomrule
    \end{tabular}
\end{table}

\end{document}